\documentclass{article} 
\usepackage{iclr2016_conference,times}
\usepackage{hyperref}
\usepackage{url}
\usepackage{amsmath,amssymb,amsthm}
\usepackage{bm}
\usepackage[utf8]{inputenc}
\usepackage[T1]{fontenc}
\usepackage{graphicx}

\title{Density Modeling of Images using a \\ Generalized Normalization Transformation}

\author{Johannes Ballé, Valero Laparra \& Eero P. Simoncelli \thanks{EPS is also affiliated with the Courant Institute of Mathematical Sciences at NYU; VL is also affiliated with the University of València, Spain.} \\
Center for Neural Science\\
New York University\\
New York, NY 10004, USA \\
\texttt{\{johannes.balle,valero,eero.simoncelli\}@nyu.edu}
}

\DeclareMathOperator{\sgn}{sgn}

\newcommand{\E}{\operatorname{\mathbb E}}
\newcommand{\R}{\operatorname{\mathbb R}}
\newcommand{\D}{\;\mathrm{d}}
\newcommand{\T}{\top}

\allowdisplaybreaks[2]

\graphicspath{{figures/}}

\iclrfinalcopy 

\begin{document}

\maketitle

\begin{abstract}
We introduce a parametric nonlinear transformation that is well-suited for Gaussianizing data from natural images. The data are linearly transformed, and each component is then normalized by a pooled activity measure, computed by exponentiating a weighted sum of rectified and exponentiated components and a constant.  We optimize the parameters of the full transformation (linear transform, exponents, weights, constant) over a database of natural images, directly minimizing the negentropy of the responses. The optimized transformation substantially Gaussianizes the data, achieving a significantly smaller mutual information between transformed components than alternative methods including ICA and radial Gaussianization. The transformation is differentiable and can be efficiently inverted, and thus induces a density model on images. We show that samples of this model are visually similar to samples of natural image patches.  We demonstrate the use of the model as a prior probability density that can be used to remove additive noise. Finally, we show that the transformation can be cascaded, with each layer optimized using the same Gaussianization objective, thus offering an unsupervised method of optimizing a deep network architecture.
\end{abstract}

\section{Introduction}
The learning of representations for classification of complex patterns has experienced an impressive series of successes in recent years.  But these results have relied heavily on large labeled data sets, leaving open the question of whether such representations can be learned directly from observed examples, without supervision.
Density estimation is the mother of all unsupervised learning problems.
A direct approach to this problem involves fitting a probability density model, either drawn from a parametric family, or composed as a nonparametric superposition of kernels, to the data. An indirect alternative, which can offer access to different families of densities, and in some cases an easier optimization problem, is to seek an invertible and differentiable parametric function $\bm y = g(\bm x; \bm \theta)$ that best maps the data onto a fixed target density $p_{\bm y}(\bm y)$.  The inverse image of this target density then provides a density model for the input space.

Many unsupervised learning methods may be interpreted in this context.
As a simple example, consider principal component analysis \citep[PCA;][]{Jo02}: we might fix $p_{\bm y}$ as a multivariate standard normal and think of PCA as either a linear whitening transformation, or as a density model $p_{\bm x}$ describing the data as a normal distribution with arbitrary covariance.
Independent component analysis \citep[ICA;][]{Ca03} can be cast in the same framework: In this case, the data $\bm x$ is modeled as a linear combination of independent heavy-tailed sources. We may fix $g$ to be linear and $p_{\bm y} = \prod_i p_{y_i}$ to be a product of independent marginal densities of unknown form. Alternatively, we can apply nonparametric nonlinearities to the marginals of the linearly transformed data so as to Gaussianize them (i.e., histogram equalization).  For this combined ICA-marginal-Gaussianization (ICA-MG) operation, $p_{\bm y}$ is again standard normal, and the transformation is a composition of a linear transform and marginal nonlinearities.
Another model that aims for the same outcome is radial Gaussianization \citep[RG;][]{LySi09a,SiBe10}, in which $g$ is the composition of a linear transformation and a {\em radial} (i.e., operating on the vector length) Gaussianizing nonlinearity. The induced density model is the family of elliptically symmetric distributions.

The notion of optimizing a transformation so as to achieve desired statistical properties at the output is central to theories of efficient sensory coding in neurobiology \citep{Ba61,Ru94,RiBoBi95,BeSe97,ScSi01}, and also lends itself naturally to the design of cascaded representations such as deep neural networks.
Specifically, variants of ICA-MG transformations have been applied in iterative cascades to learn densities
\citep{FrStSc84,ChGo00,LaCaMa11}.
Each stage seeks a linear transformation that produces the “least Gaussian” marginal directions, and then Gaussianizes these using nonparametric scalar nonlinear transformations.
In principle, this series of transformations can be shown to converge for any data density. However, the generality of these models is also their weakness: implementing the marginal nonlinearities in a non-parametric way makes the model prone to error and requires large amounts of data. In addition, since the nonlinearities operate only on marginals, convergence can be slow, requiring a lengthy sequence of transformations (i.e., a very deep network).

To address these shortcomings, we develop a  joint transformation that is highly effective in Gaussianizing local patches of natural images. The transformation is a generalization of {\em divisive normalization}, a form of local gain control first introduced as a means of modeling nonlinear properties of cortical neurons \citep{He92}, in which linear responses are divided by pooled responses of their rectified neighbors.  Variants of divisive normalization have been found to reduce dependencies when applied to natural images or sounds and to produce approximately Gaussian responses \citep{Ru94,ScSi01,MaLa10}. Simple forms of divisive normalization have been shown to offer improvements in recognition  performance of deep neural networks \citep{JaKaRaLe09}. But  the Gaussianity of these representations has not been carefully optimized, and typical forms of normalization do not succeed in capturing all forms of dependency found in natural images \citep{Ly10,SiBe13}.

In this paper, we define a generalized divisive normalization (GDN) transform that includes parametric forms of both ICA-MG and RG as special cases. We solve for the parameters of the transform by optimizing an unsupervised learning objective for the non-Gaussianity of the transformed data. The transformation is continuous and differentiable, and we present an effective method of inverting it.  We demonstrate that the resulting GDN transform provides a significantly better model for natural photographic images than either ICA-MG or RG.  Specifically, we show that GDN provides a better fit to the pairwise statistics of local filter responses, that it generates more natural samples of image patches, and that it produces better results when used as a prior for image processing problems such as denoising.  Finally, we show that a two-stage cascade of GDN transformations offers additional improvements in capturing image statistics, laying the groundwork for its use as a general tool for unsupervised learning of deep networks.

\section{Parametric Gaussianization}
Given a parametric family of transformations $\bm y = g(\bm x; \bm\theta)$, we wish to select parameters $\bm \theta$ so as to transform the input vector $\bm x$ into a standard normal random vector (i.e., zero mean, identity covariance matrix).
For a differentiable transformation, the input and output densities are related by:
\begin{equation}
p_{\bm x}(\bm x) = \left| \frac {\partial g(\bm x; \bm\theta)} {\partial \bm x} \right| \, p_{\bm y}\bigl(g(\bm x; \bm \theta)\bigr) ,
\label{eq:p_x}
\end{equation}
where $|\cdot|$ denotes the absolute value of the matrix determinant. If $p_{\bm y}$ is the standard normal distribution (denoted $\mathcal N$), the shape of $p_{\bm x}$ is determined solely by the transformation.  Thus, $g$ \emph{induces} a density model on $\bm x$, specified by the parameters $\bm \theta$.

Given $p_{\bm x}$, or data drawn from it, the density estimation problem can be solved by minimizing the Kullback--Leibler (KL) divergence between the transformed density and the standard normal, known as the {\em negentropy}:
\begin{equation}
J(p_{\bm y}) = \E_{\bm y} \Bigl( \log p_{\bm y}(\bm y) - \log \mathcal N( \bm y ) \Bigr) = \E_{\bm x} \Bigl( \log p_{\bm x}(\bm x) - \log\Bigl| \tfrac {\partial g(\bm x;\bm\theta)} {\partial \bm x} \Bigr| - \log \mathcal N(g(\bm x; \bm\theta)) \Bigr),
\label{eq:kl}
\end{equation}
where we have rewritten the standard definition (an expected value over ${\bm y}$) as an expectation over ${\bm x}$ (see appendix).
Differentiating with respect to the parameter vector $\bm\theta$ yields:
\begin{equation}
\frac {\partial J(p_{\bm y})} {\partial \bm \theta} =
\E_{\bm x} \left(
  - \sum_{ij} \left[ \frac {\partial g(\bm x, \bm \theta)} {\partial \bm x} \right]^{-\T}_{ij} \frac {\partial^2 g_i(\bm x, \bm \theta)} {\partial x_j \partial \bm \theta} +
\sum_i g_i(\bm x, \bm \theta) \frac {\partial g_i(\bm x, \bm \theta)} {\partial \bm \theta} \right),
\label{eq:gradient}
\end{equation}
where the expectation can be evaluated by summing over data samples, allowing the model to be fit using stochastic gradient descent. It can be shown that this optimization is equivalent to maximizing the log likelihood of the induced density model.

Note that, while optimization is feasible, measuring success in terms of the actual KL divergence in eq.~\eqref{eq:kl} is difficult in practice, as it requires evaluating the entropy of $p_{\bm x}$. Instead, we can monitor the difference in negentropy between the input and output densities:
\begin{equation}
\Delta J \equiv J(p_{\bm y}) - J(p_{\bm x}) = \E_{\bm x} \Bigl( \tfrac 1 2 \bigl\| \bm y \bigr\|_2^2 - \log\Bigl| \tfrac {\partial \bm y} {\partial \bm x} \Bigr| - \tfrac 1 2 \bigl\| \bm x \bigr\|_2^2 \Bigr).
\label{eq:delta_J}
\end{equation}
This quantity provides a measure of how much more Gaussian the data become as a result of the transformation $g(\bm x; \bm\theta)$.

\section{Divisive normalization transformations}
Divisive normalization, a  form of gain control in which responses are divided by pooled activity of neighbors, has become a standard model for describing the nonlinear properties of sensory neurons \citep{CaHe12}. A commonly used form for this transformation is:
\begin{equation*}
y_i = \gamma \frac {x_i^\alpha} {\beta^\alpha + \sum_j x_j^\alpha},
\label{eq:dn_simple}
\end{equation*}
where $\bm \theta = \{ \alpha, \beta, \gamma \}$ are parameters. Loosely speaking, the transformation adjusts responses to lie within a desired operating range, while maintaining their relative values.
A weighted form of normalization (with exponents fixed at $\alpha=2$) was introduced in \citep{ScSi01}, and shown to produce approximately Gaussian responses with greatly reduced dependencies. The weights were optimized over a collection of photographic images so as to maximize the likelihood of responses under a Gaussian model.  Normalization has also been derived as an inference method for a Gaussian scale mixture (GSM) model for wavelet coefficients of natural images \citep{WaSi00}. This model factorizes local groups of coefficients into a Gaussian vector and a positive-valued scalar. In a specific instance of the model, the optimal estimator for the Gaussian vector (after decorrelation) can be shown to be a modified form of divisive normalization that uses a weighted $L_2$-norm \citep{LySi08}:
\begin{equation*}
y_i = \frac {x_i} {\bigl(\beta^2 + \sum_j \gamma_j x_j^2\bigr)^{\frac 1 2}}.
\end{equation*}
However, the above instances of divisive normalization have only been shown to be effective when applied to spatially local groups of filter responses. In what follows, we introduce a more general form, with better Gaussianization capabilities that extend to to more distant responses, as well as those arising from distinct filters.

\subsection{Proposed generalized divisive normalization (GDN) transform}
We define a vector-valued parametric transformation as a composition of a linear transformation followed by a generalized form of divisive normalization:
\begin{align}
\label{eq:g}
\bm y = g(\bm x; \bm \theta)
  && \text{s.t.} && y_i &= \frac {z_i} {\bigl(\beta_i + \sum_j \gamma_{ij}|z_j|^{\alpha_{ij}}\bigr)^{\varepsilon_i}} \notag \\
  && \text{and} && \bm z &= \bm H \bm x.
\end{align}
The full parameter vector $\bm \theta$ consists of the vectors
$\bm \beta$ and $\bm \varepsilon$,
as well as the matrices $\bm H$, $\bm \alpha$, and $\bm \gamma$, for a total of $2N + 3N^2$ parameters (where $N$ is the dimensionality of the input space). We refer to this transformation as {\em generalized divisive normalization} (GDN), since it generalizes several previous models. Specifically:
\begin{itemize}
\item Choosing $\varepsilon_i \equiv 1$,  $\alpha_{ij} \equiv 1$, and $\gamma_{ij} \equiv 1$ yields the classic form of the divisive normalization transformation \citep{CaHe12}, with exponents set to 1.
\item Choosing $\bm \gamma$ to be diagonal eliminates the cross terms in the normalization pool, and the model is then a particular form of ICA-MG, or the first iteration of the Gaussianization algorithms described in \cite{ChGo00} or \cite{LaCaMa11}: a linear “unmixing” transform, followed by a pointwise, Gaussianizing nonlinearity.
\item Choosing $\alpha_{ij} \equiv 2$
and setting all elements of $\bm \beta$, $\bm \varepsilon$, and $\bm \gamma$ identical, the transformation assumes a radial form:
\begin{equation*}
\bm y = \frac {\bm z} {\bigl(\beta + \gamma \sum_j z_j^2\bigr)^{\varepsilon}} = \frac {\bm z} {\|\bm z\|_2} \, g_2\bigl( \| \bm z \|_2 \bigr)
\end{equation*}
where $g_2(r) = r /(\beta + \gamma r^2)^{\varepsilon}$ is a scalar-valued transformation on the radial component of $\bm z$, ensuring that the normalization operation preserves the vector direction of $\bm z$. If, in addition, $\bm H$ is a whitening transformation such as ZCA \citep{BeSe97}, the overall transformation is a form of RG \citep{LySi09a}.
\item More generally, if we allow exponents $\alpha_{ij} \equiv p$, the induced distribution is an $L_p$-symmetric distribution, a family which has been shown to capture various statistical properties of natural images \citep{SiBe10}. The corresponding transformation on the $L_p$-radius is given by $g_p(r) = r /(\beta + \gamma r^p)^{\varepsilon}$.
\item Another special case of interest arises when partitioning the space into distinct, spherically symmetric subspaces, with the $k$th subspace comprising the set of vector indices $S_k$. Choosing $\alpha_{ij} \equiv 2$, $\beta_i = \beta_k'$, $\varepsilon_i = \varepsilon_k'$, and $\gamma_{ij} = \gamma_k'$, all for $i,j \in S_k$ (and $\gamma_{ij} = 0$ if $i$ or $j$ is not in the same set), the nonlinear transformation can be written as
\begin{equation*}
y_i = \frac {z_i} {\bigl(\beta_k' + \gamma_k' \sum_{j \in S_k} z_j^2\bigr)^{\varepsilon_k'}},
\end{equation*}
where $k$ is chosen such that $i \in S_k$. This is the Independent Subspace Analysis model \citep[ISA;][]{HyHo00}, expressed as a Gaussianizing transformation.
\end{itemize}
The topographic ICA model \citep[TICA;][]{HyHoIn01} and the model presented in \cite{KoHy10} are generalizations of ISA that are related to our model, but have more constrained nonlinearities. They are formulated directly as density models, which makes them difficult to normalize. For this reason, the authors must optimize approximated likelihood or use score matching \citep{Hy05} to fit these models.

\subsection{Well-definedness and Invertibility}
For the density function in eq.~\eqref{eq:p_x} to be well defined, we require the transformation in eq.~\eqref{eq:g} to be continuous and invertible. For the linear portion of the transformation, we need only ensure that the matrix $\bm H$ is non-singular. For the normalization portion, consider the partial derivatives:
\begin{equation}
\label{eq:dydz}
\frac {\partial y_i} {\partial z_k} = \frac {\delta_{ik}} {\bigl(\beta_i + \sum_j \gamma_{ij} |z_j|^{\alpha_{ij}}\bigr)^{\varepsilon_i}} - \frac {\alpha_{ik} \gamma_{ik} \varepsilon_i z_i |z_k|^{\alpha_{ik} - 1} \sgn(z_k)} {\bigl(\beta_i + \sum_j \gamma_{ij} |z_j|^{\alpha_{ij}}\bigr)^{\varepsilon_i+1}}
\end{equation}
To ensure continuity, we require all partial derivatives to be finite for all $\bm z \in \R^N$. More specifically, we require all exponents in eq.~\eqref{eq:dydz} to be non-negative, as well as the parenthesized expression in the denominator to be positive.

It can be shown that the normalization part of the transformation is invertible if the Jacobian matrix containing the partial derivatives in eq.~\eqref{eq:dydz} is positive definite everywhere (see appendix). In all practical cases, we observed this to be the case, but expressing this precisely as a condition on the parameters is difficult. A necessary (but generally not sufficient) condition for invertibility can be established as follows. First, note that, as the denominator is positive, each vector $\bm z$ is mapped to a vector $\bm y$ in the same orthant. The cardinal axes of $\bm z$ are mapped to themselves, and for this one-dimensional mapping to be continuous and invertible, it must be monotonic.
Along the cardinal axes, the following bound holds:
\begin{equation*}
|y_i| = \frac {|z_i|} {\bigl(\beta_i + \gamma_{ii}|z_i|^{\alpha_{ij}}\bigr)^{\varepsilon_i}}
\le \frac {|z_i|} {\gamma_{ii}^{\varepsilon_i}|z_i|^{\alpha_{ii}\varepsilon_i}}
= \gamma_{ii}^{-\varepsilon_i} |z_i|^{1-\alpha_{ii}\varepsilon_i}.
\end{equation*}
For the magnitude of $y_i$ to grow monotonically with $|z_i|$, the exponent $1-\alpha_{ii}\varepsilon_i$ must be positive.

In summary, the constraints we enforce during the optimization are $\alpha_{ij} \geq 1$, $\beta_i > 0$, $\gamma_{ij} \geq 0$, and $0 \leq \varepsilon_i \leq \alpha_{ii}^{-1}$. We initialize the parameters such that $\frac{\partial \bm y}{\partial \bm z}$ is positive definite everywhere (for example, by letting $\bm \gamma$ be diagonal, such that the Jacobian is diagonal, the transformation is separable, and the necessary constraint on the cardinal axes becomes sufficient).

Suppose that during the course of optimization, the matrix should cease to be positive definite. Following a continuous path, the matrix must then become singular at some point, because going from positive definite to indefinite or negative definite would require at least one of the eigenvalues to change signs. However, the optimization objective heavily penalizes singularity: The term $-\log| \frac {\partial \bm y} {\partial \bm x} |$ in the objective (which separates into $-\log| \frac {\partial \bm y} {\partial \bm z} |$ and $- \log| \frac {\partial \bm z} {\partial \bm x} |$) grows out of bounds as the determinant of the Jacobian approaches zero. Therefore, given a sensible initialization and a sufficiently small step size, it is highly unlikely that the Jacobian should cease to be positive definite during the optimization, and we haven't observed this in practice.

Finally, we find that the GDN transformation can be efficiently inverted using a fixed point iteration:
\begin{align*}
z_i^{(0)} &= \sgn( y_i ) \, \bigl( \gamma_{ii}^{\varepsilon_i} |y_i| \bigr)^{\frac 1 {1 - \alpha_{ii} \varepsilon_i}} \\
z_i^{(n+1)} &= \Bigl(\beta_i + \sum_j \gamma_{ij} \bigl|z_j^{(n)}\bigr|^{\alpha_{ij}}\Bigr)^{\varepsilon_i} y_i .
\end{align*}
Other iterative inverse solutions have been proposed for this purpose \citep{MaEpNaSi06,LySi08}, but these only apply to special cases of the form in eq.~\eqref{eq:g}.

\section{Experiments}
The model was optimized to capture the distribution of image data using stochastic descent of the gradient expressed in eq.~\eqref{eq:gradient}.  We then conducted a series of experiments to assess the validity of the fitted model for natural images.

\subsection{Joint density of pairs of wavelet coefficients}
\begin{figure}[p]
\centering
\includegraphics*[scale=.66,trim=5 5 5 5]{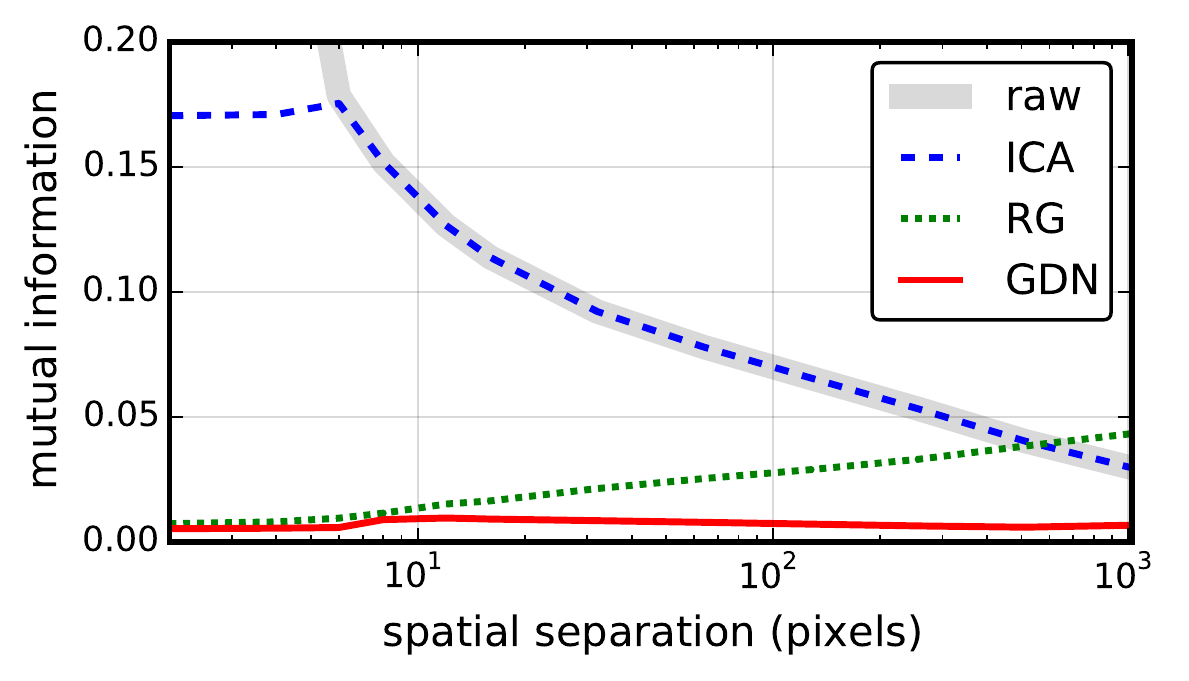}
\caption{Mutual information in pairs of wavelet coefficients after various transformations, plotted as a function of the spatial separation between the coefficients.}
\label{fig:mi}
\end{figure}

We examined the pairwise statistics of model responses, both for our GDN model, as well as the ICA model and the RG model.  First, we computed the responses of an oriented filter (specifically, we used a subband of the steerable pyramid \citep{SiFr95}) to images taken from the van Hateren dataset \citep{HaSc98} and extracted pairs of coefficients within subbands at different spatial offsets up to $d=1024$. We then transformed these two-dimensional datasets using ICA, RG, and GDN. Figure~\ref{fig:mi} (modelled after figure 4 of \citet{LySi09a}) shows the mutual information in the transformed data (note that, in our case, mutual information is related to the negentropy by an additive constant.) For very small distances, a linear ICA transformation reduces some of the dependencies in the raw data. However, for larger distances, a linear transformation is not sufficient to eliminate the dependencies between the coefficients, and the mutual information of the ICA-transformed data is identical to that of the raw data. An elliptically symmetric model is good for modeling the density when the distance is small, and the RG transform reduces the mutual information to neglegible levels. However, the fit worsens as the distance increases. As described in \citep{LySi09a}, RG can even lead to an increase in the mutual information relative to that of the raw data, as seen in the right hand side of the plot. The GDN transform, however, captures the dependencies at all separations, and consistently leaves a very small residual level of mutual information.

\begin{figure}[p]
ICA-MG\hfill%
\raisebox{-57pt}{%
\includegraphics*[width=.30\linewidth,trim=5 15 15 5]{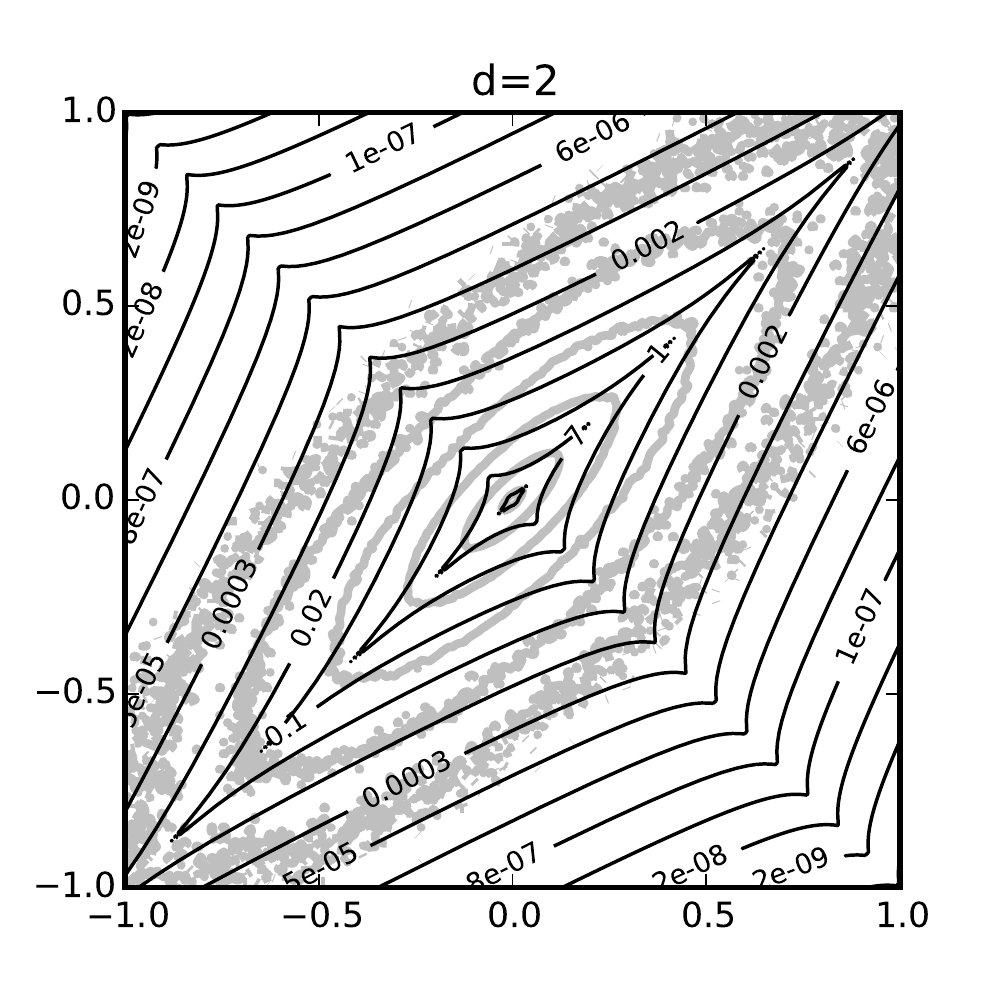}%
\includegraphics*[width=.30\linewidth,trim=5 15 15 5]{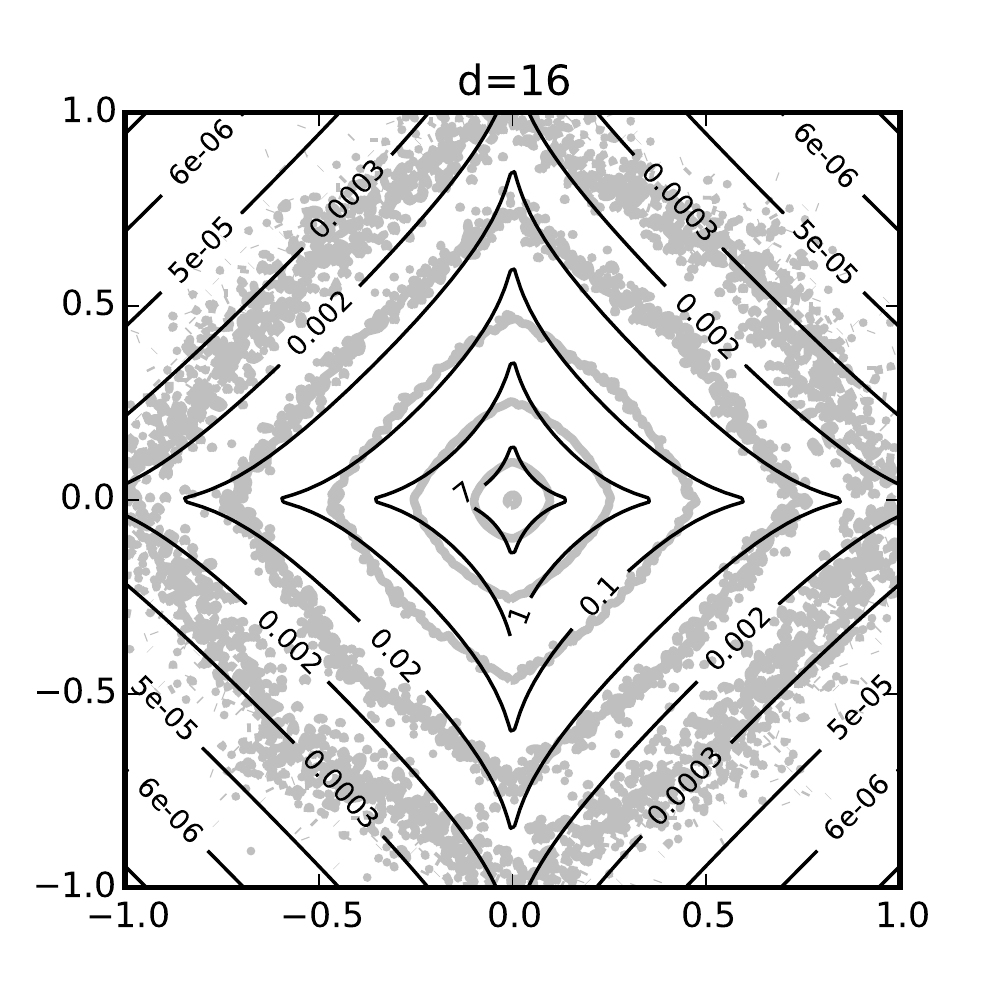}%
\includegraphics*[width=.30\linewidth,trim=5 15 15 5]{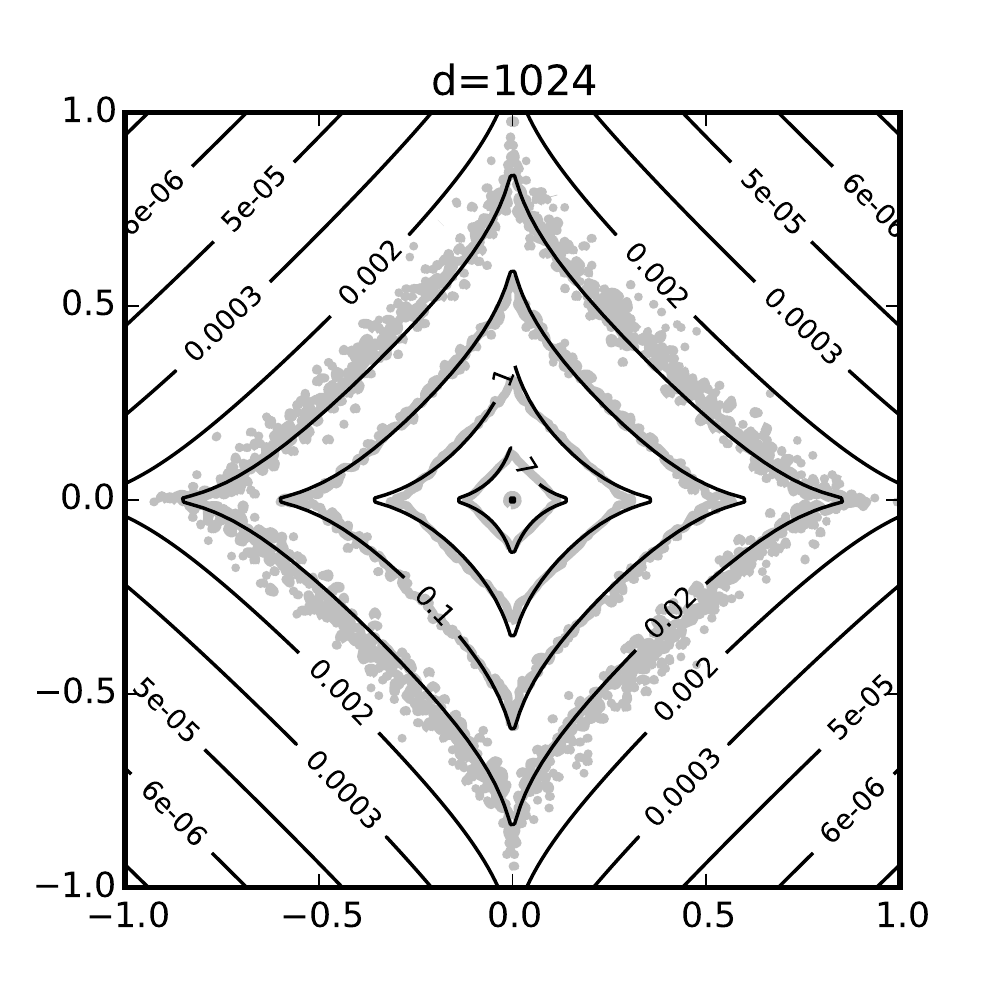}%
}\\
RG\hfill%
\raisebox{-57pt}{%
\includegraphics*[width=.30\linewidth,trim=5 15 15 5]{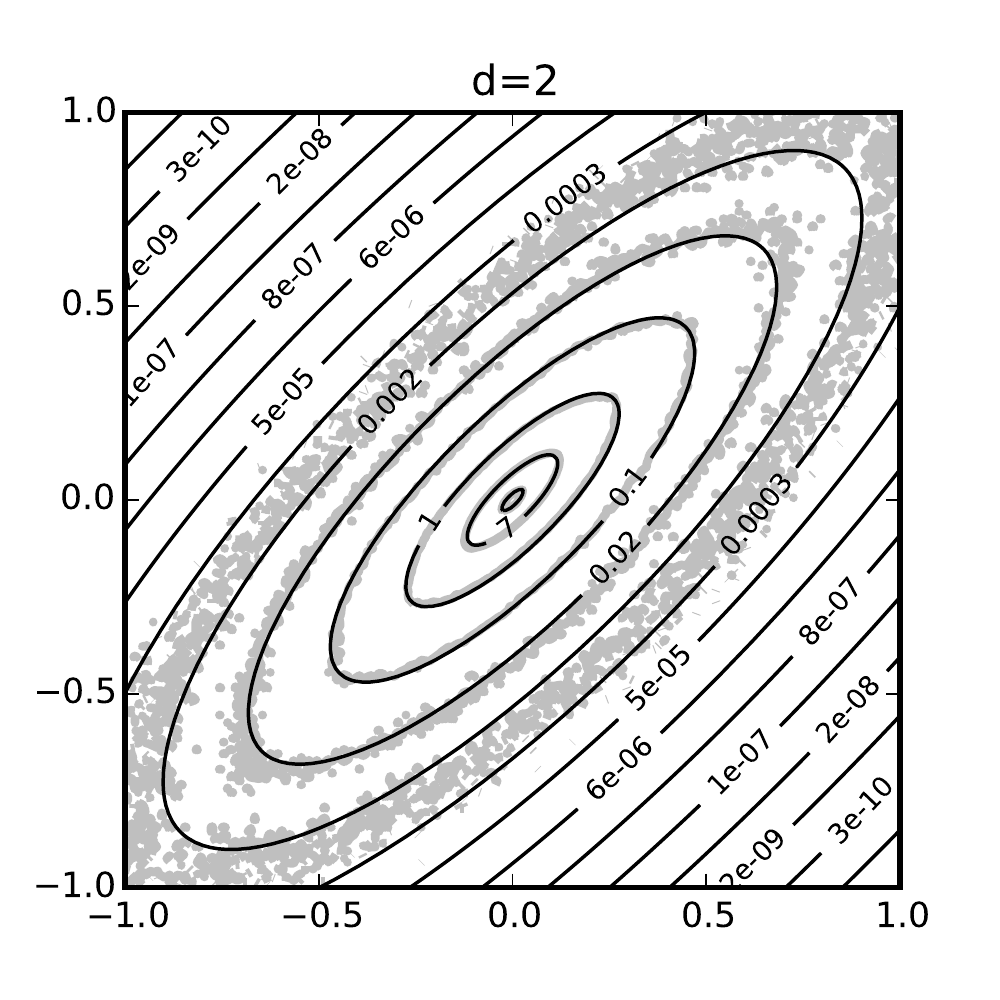}%
\includegraphics*[width=.30\linewidth,trim=5 15 15 5]{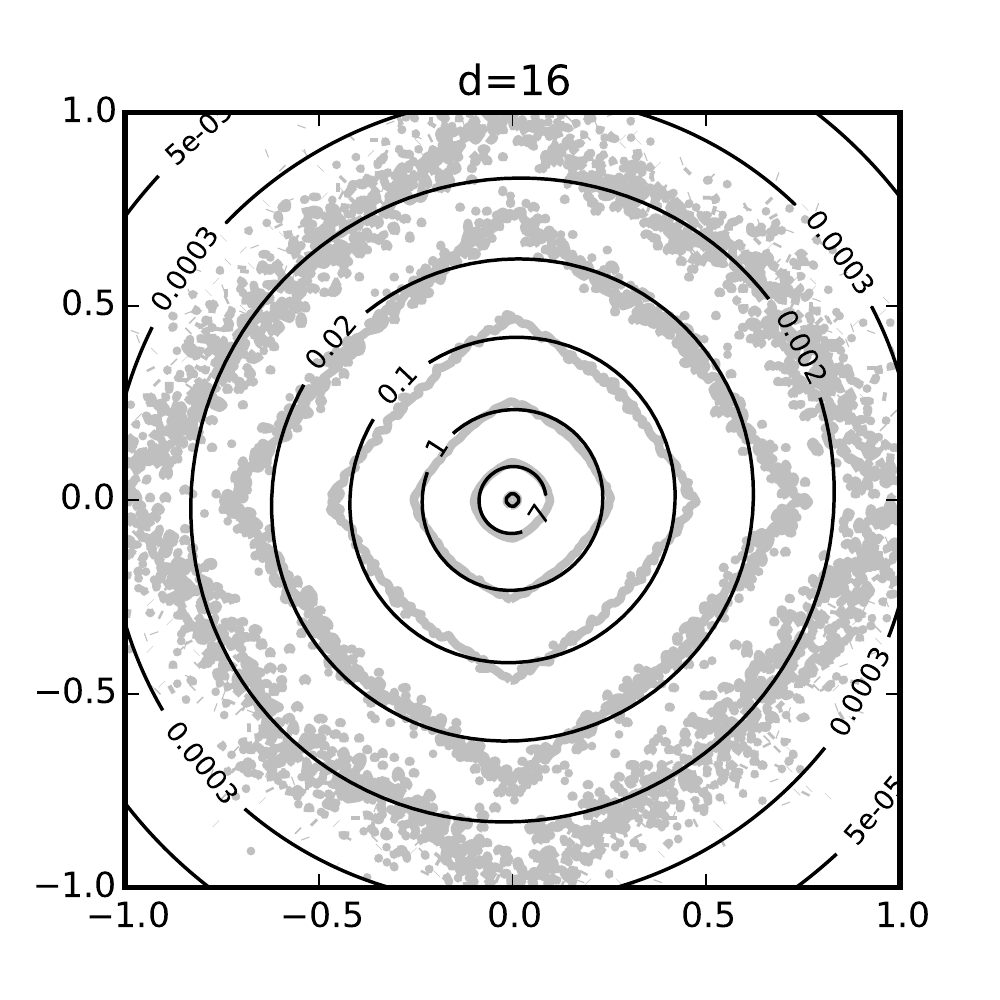}%
\includegraphics*[width=.30\linewidth,trim=5 15 15 5]{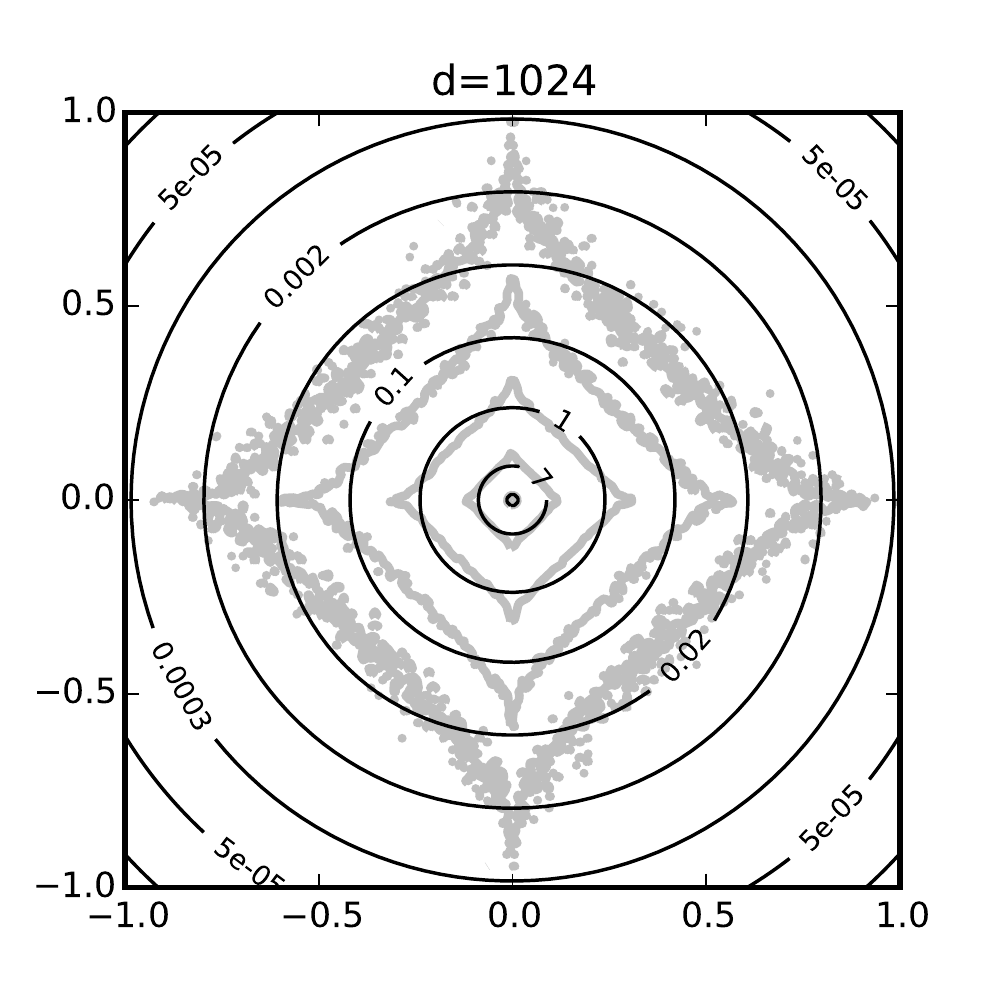}%
}\\
GDN\hfill%
\raisebox{-57pt}{%
\includegraphics*[width=.30\linewidth,trim=5 15 15 5]{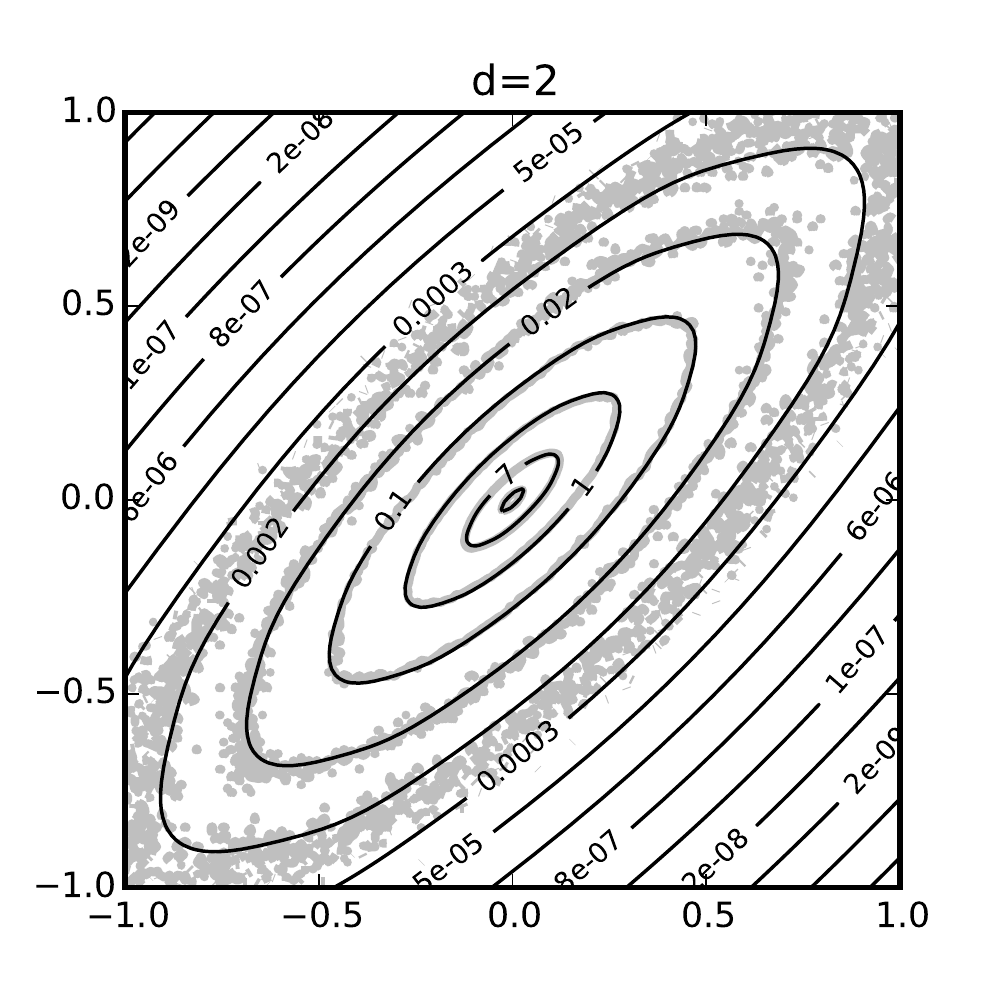}%
\includegraphics*[width=.30\linewidth,trim=5 15 15 5]{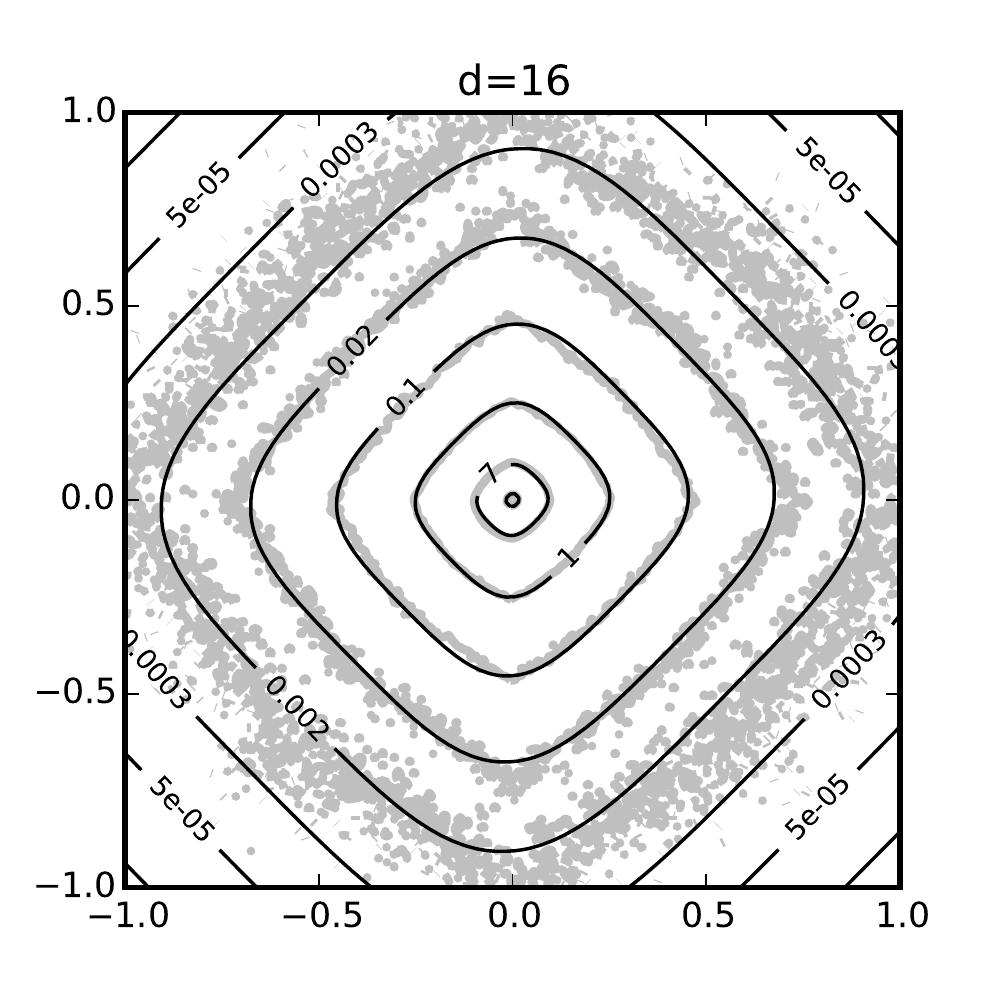}%
\includegraphics*[width=.30\linewidth,trim=5 15 15 5]{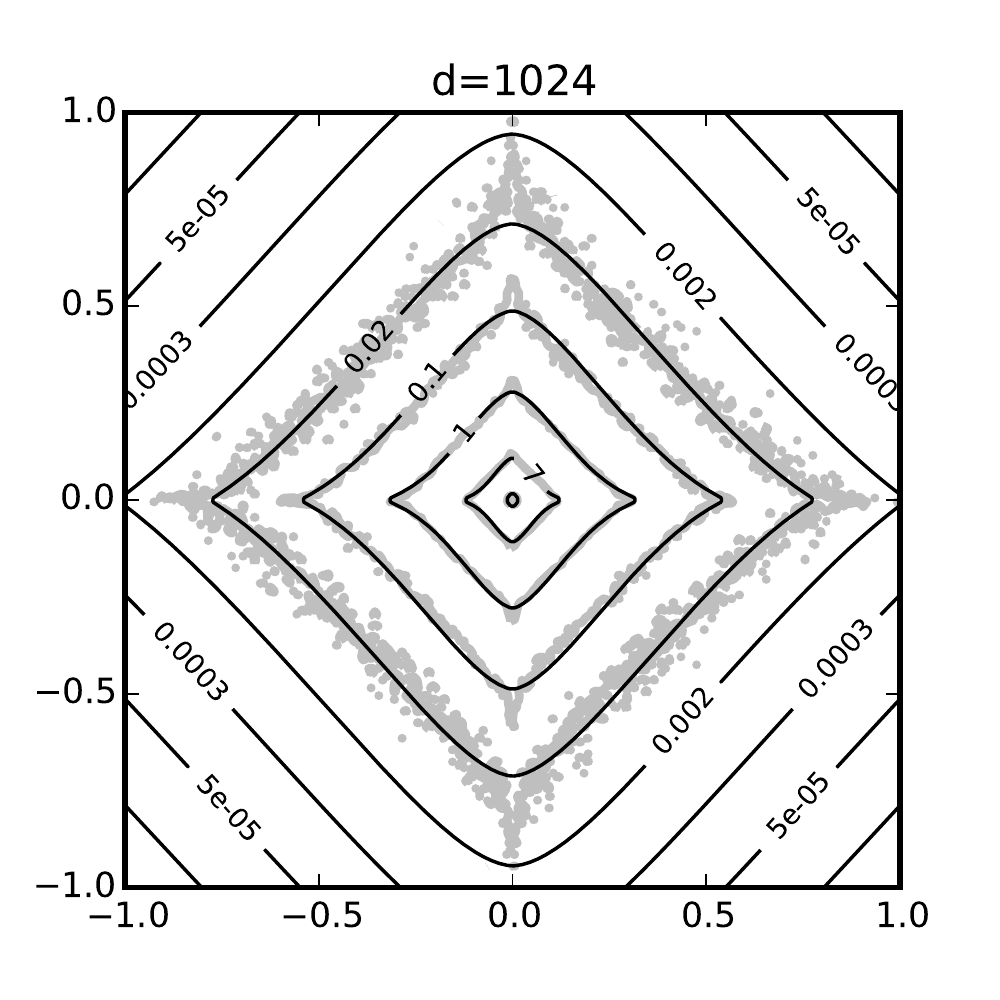}%
}\\
\caption{Contour plots of pairwise wavelet coefficient densities.  Each row corresponds to a model arising from a different transformation (ICA-MG, RG, GDN). Each column corresponds to a pair of coefficients spatially separated by distance $d$ (pixels). Gray: contour lines of histogram density estimate. Black: contour lines of densities induced by best-fitting transformations. As distance increases, the empirical density between the coefficients transitions from elliptical but correlated to separable. The RG density captures the former, and the ICA density captures the latter.  Only the GDN density has sufficient flexibility to capture the full range of behaviors.}
\label{fig:contours}
\end{figure}

In figure~\ref{fig:contours}, we compare histogram estimates of the joint wavelet coefficient densities against model-fitted densities for selected spatial offsets. The GDN fits are seen to account well for the shapes of the densities, particularly in the center of the plot, where the bulk of the data lie. Note that GDN is able to capture elliptically symmetric distributions just as well as distributions which are closer to being marginally independent, whereas RG and ICA each fail in one of these cases, respectively.

\subsection{Joint density over image patches}
We also examined model behavior when applied to vectorized $16\times 16$ blocks of pixels drawn from the Kodak set\footnote{downloaded from http://www.cipr.rpi.edu/resource/stills/kodak.html}. We used the stochastic optimization algorithm \textsc{Adam} to facilitate the optimization \citep{KiBa15} and somewhat reduced the complexity of the model by forcing $\bm \alpha$ to be constant along its columns (i.e., $\alpha_{ij}\equiv\alpha_j$). We also fitted versions of the model in which the normalization (denominator of eq.~\eqref{eq:g}) is constrained to marginal transformations (ICA-MG) or radial transformations (RG). For higher dimensional data, it is difficult to visualize the densities, so we use other measures to evaluate the effectiveness of the model:

{\bf Negentropy reduction.}
As an overall metric of model fit, we evaluated the negentropy difference $\Delta J$ given in \eqref{eq:delta_J} on the full GDN model, as well as the marginal and radial models model (ICA-MG and RG, respectively).
We find that ICA-MG and RG reduce negentropy by 2.04 and 2.11 nats per pixel, respectively, whereas GDN reduces it by 2.43 nats.

\begin{figure}[t]
\centering
\includegraphics*[scale=.66,trim=5 5 5 5]{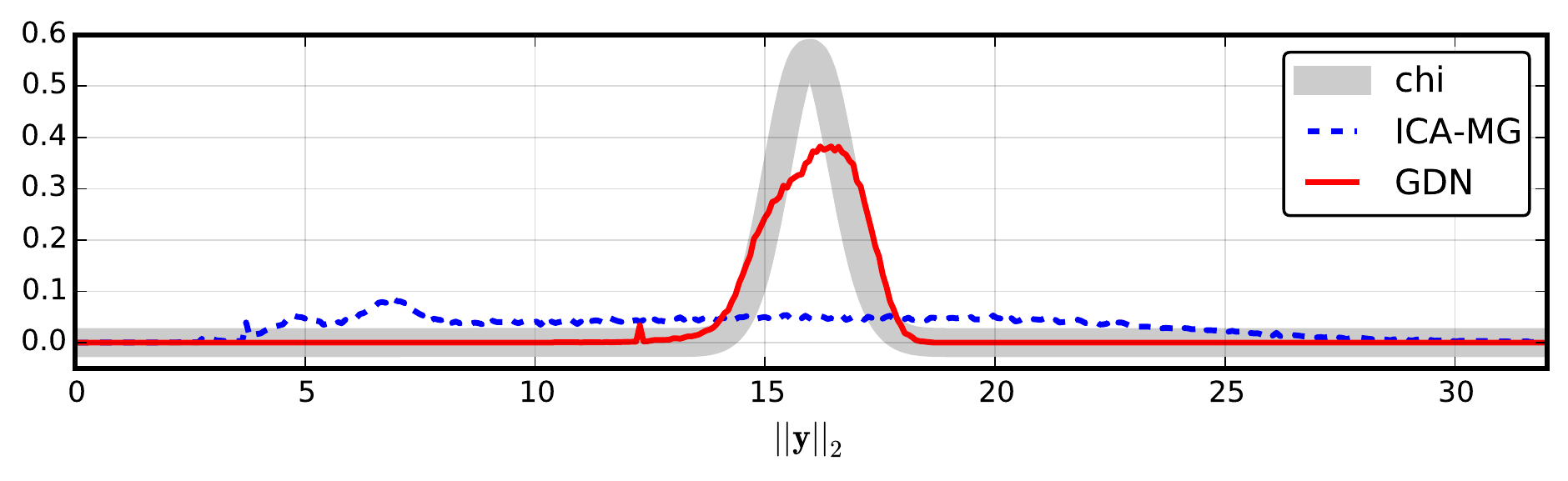} \\[1ex]
\includegraphics*[scale=.66,trim=5 5 5 5]{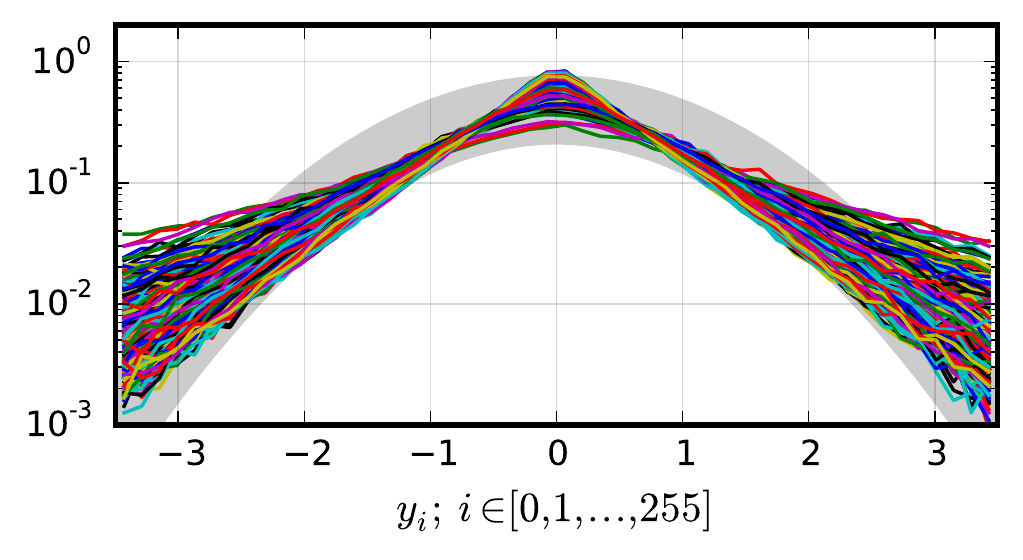}%
\hfil%
\includegraphics*[scale=.66,trim=5 5 5 5]{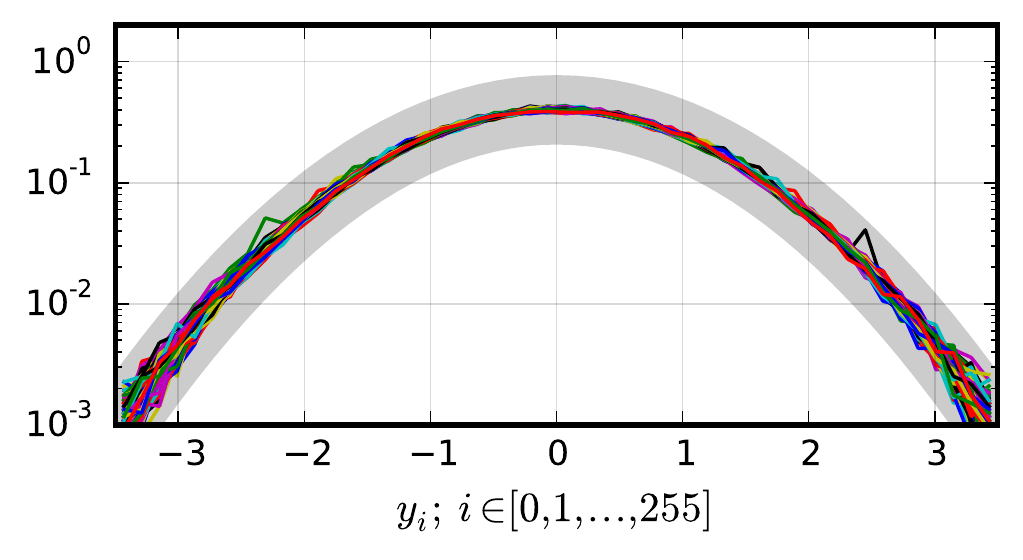}%
\caption{Histograms of transformed data. Top: Radial component for ICA-MG and GDN. Bottom left and right: Marginals for RG and GDN, respectively. Gray lines indicate the expected distributions (Chi for radial, and Gaussian for marginals).}
\label{fig:plots}
\end{figure}

{\bf Marginal/radial distributions of transformed data.}
If the transformed data is multivariate standard normal, its marginals should be standard normal, as well, and the radial component should be Chi distributed with degree 256. Figure~\ref{fig:plots} shows these distributions, in comparison to those of ICA-MG and RG.  As expected from \citep{LySi09a}, RG fails to Gaussianize the marginals, and ICA-MG fails to transform the radial component into a Chi distribution. GDN comes close to achieving both goals.

\begin{figure}[t]
\centering
\includegraphics*[width=.27\linewidth]{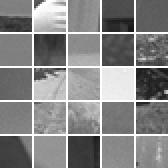}%
\hspace{3ex}%
\includegraphics*[width=.27\linewidth]{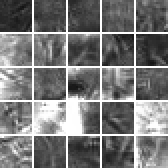}%
\hspace{3ex}%
\includegraphics*[width=.27\linewidth]{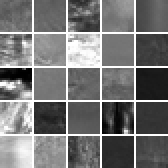}%
\caption{Sample image patches. From left to right, 25 samples drawn from: the image training set; the ICA-MG model; the GDN model.}
\label{fig:samples}
\end{figure}

{\bf Sampling.}
The density model induced by a transformation can also be visualized by examining samples drawn from a standard normal distribution that have been passed through the inverse transformation. Figure~\ref{fig:samples} compares sets of 25 image patches drawn from the GDN model, the ICA-MG model, and randomly selected from a database of images.  GDN notably captures two features of natural images: First, a substantial fraction of the samples are constant or nearly so (as in the natural images, which include patches of sky or untextured surfaces). Second, in the cases with more activity, the samples contain sparse “organic” structures (although less so than those drawn from the natural images). In comparison, the samples from the ICA-MG model are more jumbled, and filled with random mixtures of oriented elements.

{\bf Denoising.}
The negentropy provides a particular metric for assessing the quality of our results, but it need not agree with other measures \citep{ThOoBe15}.  Another test of a probability model comes from using it as a prior in a Bayesian inference problem.  The most basic example is that of removing additive Gaussian noise.  For GDN, we use the empirical Bayes solution of \cite{Mi61}, which expresses the least-squares optimal solution {\em directly} as a function of the distribution of the noisy data:
\begin{equation}
\bm{\hat x} = \bm{\tilde x} + \sigma^2 \, \nabla \log p_{\bm {\tilde x}} (\bm {\tilde x}),
\label{eq:stein}
\end{equation}
where $\bm{\tilde x}$ is the noisy observation, $p_{\bm {\tilde x}}$ is the density of the noisy data, $\sigma^2$ is the noise variance, and $\bm{\hat x}$ is the optimal estimate. Note that, counterintuitively, this expression does not refer directly to the prior density, but it is nevertheless exactly equivalent to the Bayesian least squares solution \citep{RaSi11}.  Although the GDN model was developed for modeling the distribution of clean image data, we use it here to estimate the distribution of the {\em noisy} image data. We find that, since the noisy density is a Gaussian-smoothed version of the original density, the model fits the data well (results not shown).

\begin{figure}[t]
\centering\footnotesize
\includegraphics*[width=.48\linewidth]{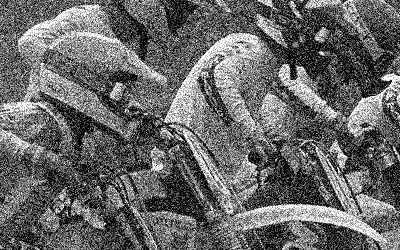}%
\hspace{3ex}%
\includegraphics*[width=.48\linewidth]{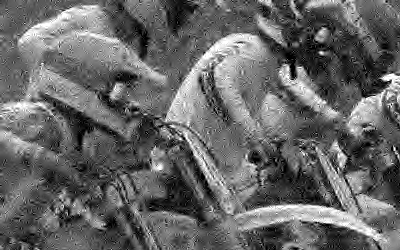}%

\makebox[.48\linewidth]{noisy}%
\hspace{3ex}%
\makebox[.48\linewidth]{marginal: PSNR 20.6, SSIM 0.68}%

\vspace{2ex}

\includegraphics*[width=.48\linewidth]{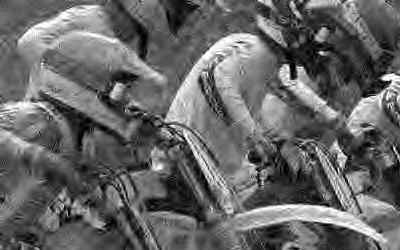}%
\hspace{3ex}%
\includegraphics*[width=.48\linewidth]{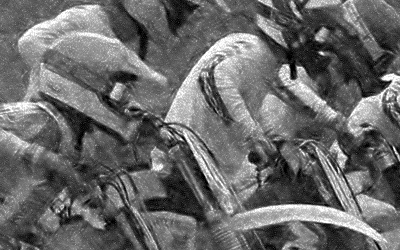}%

\makebox[.48\linewidth]{GSM: PSNR 22.4, SSIM 0.75}%
\hspace{3ex}%
\makebox[.48\linewidth]{GDN: PSNR 22.6, SSIM 0.78}%
\caption{Bayesian least squares denoising using different prior models. Top: noise-corrupted original; denoised with marginal model in an orthonormal wavelet decomposition. Bottom: denoised with GSM model in an orthonormal wavelet decomposition; denoised with GDN-induced density model. Below each image, errors against the original image are quantified with PSNR in dB, and the perceptual SSIM metric (for both measures, bigger is better).}
\label{fig:denoising}
\end{figure}

For comparison, we implemented two denoising methods that operate on orthogonal wavelet coefficients, one assuming a marginal model \citep{FiNo01}, and the other an elliptically symmetric Gaussian scale mixture (GSM) model \citep{PoStWaSi03}. Since the GDN model is applied to $16\times 16$ patches of pixels and is restricted to a complete (i.e., square matrix) linear transformation, we restrict the wavelet transform employed in the other two models to be orthogonal, and to include three scales. We also report numerical scores: the peak signal to noise ratio (PSNR), and the structural similarity index \citep[SSIM;][]{WaBoShSi04} which provides a measure of perceptual quality. Fig.~\ref{fig:denoising} shows the denoising results. Both marginal and spherical models produce results with strong artifacts resembling the basis functions of the respective linear transform. The GDN solution has artifacts that are less perceptually noticeable, while at the same time leaving a larger amount of background noise.

{\bf Average model likelihood.}
To further assess how our model compares to existing work, we trained the model on image patches of $8\times 8$ pixels from the BSDS300 dataset which had the patch mean removed \citep[see][left column of table 1]{ThBe15}. We followed the same evaluation procedures as in that reference, and measured a cross-validated average log likelihood of 126.8 nats for ICA-MG and 151.5 nats for GDN, similar to the values reported there for the RIDE model, but worse than the best-performing MCGSM and RNADE models. On the other hand, GDN achieves a model likelihood of 3.47 bits/pixel for image patches of $8\times8$ pixels without mean removal, which is essentially equal to the best reported performance in the middle column of table 1 (ibid.).\footnote{Note, however, that the middle column in table 1 of \cite{ThBe15} was generated under the assumption that the patch mean is statistically independent from the rest of the data, which artificially impedes the performance of the reported models.}

\subsection{Two-stage cascaded model}
In the previous section, we show that the GDN transformation works well on local patches of images. However, this cannot capture statistical dependencies over larger spatial distances (i.e., across adjacent patches). One way of achieving this is to cascade Gaussianizing transformations \citep{ChGo00,LaCaMa11}. In previous implementations of such cascades, each stage of the transformation consists of a linear transformation (to rotate the previous responses, exposing additional non-Gaussian directions) and a Gaussianizing nonlinear transformation applied to the marginals. We have implemented a cascade based on GDN that benefits from two innovations.  First, by jointly Gaussianizing groups of coefficients (rather than transforming each one independently), GDN achieves a much more significant reduction in negentropy than MG (see Figure \ref{fig:mi}), thereby reducing the total number of stages that would be needed to fully Gaussianize the data. Second, we replace the ICA rotations with convolutional ICA \citep[CICA;][]{BaSi14}. This is a better solution than either partitioning the image into non-overlapping blocks (which produces artifacts at block boundaries) or simply increasing the size of the transformation, which would require a much larger number of parameters for the linear transform than a convolutional solution (which allows ``weight sharing'' \citep{LeMaBoDeHe90}).

\begin{figure}[t]
\centering
\includegraphics*[scale=.66,trim=5 5 5 5]{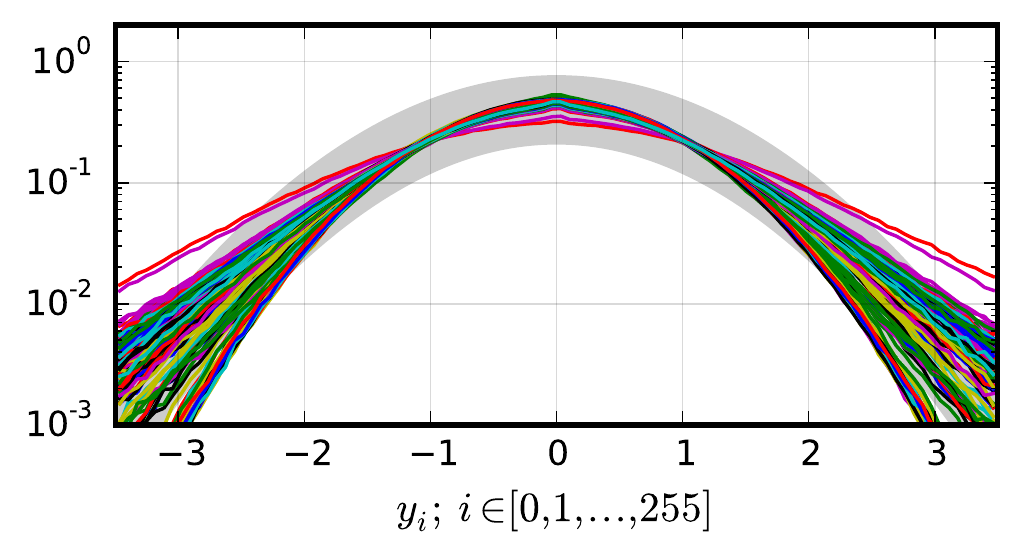}%
\hfil%
\includegraphics*[scale=.66,trim=5 5 5 5]{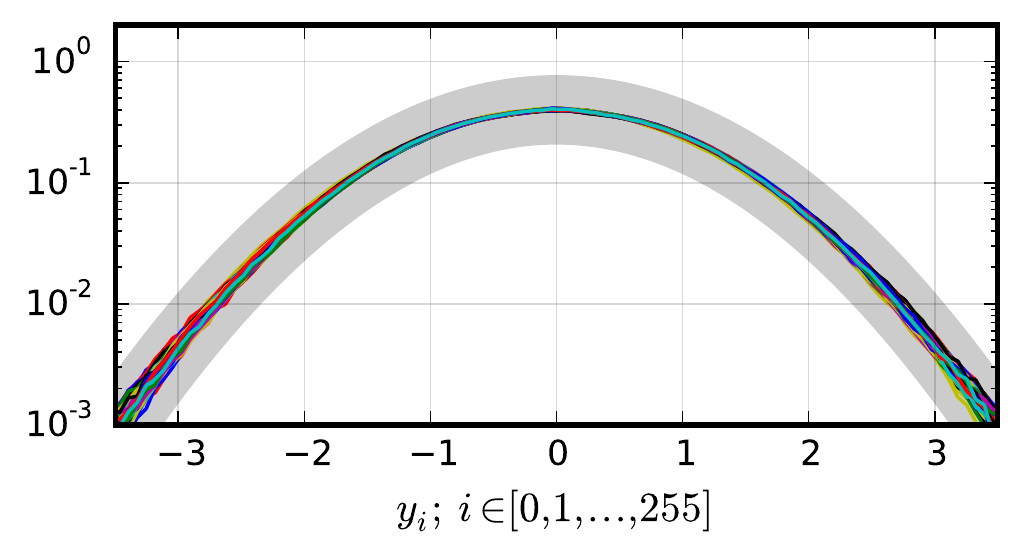} \\
\vspace*{-1ex}
\caption{Marginal histograms of two-stage model responses. Left: marginals of $256$ features, obtained by performing linear CICA on first stage GDN responses. Right: marginals after the second stage GDN. The thick gray line corresponds to a Gaussian distribution.}
\label{fig:multilayer}
\end{figure}

The central question that determines effectiveness of a multi-layer model based on the above ingredients is whether the parametric form of the normalization is suitable for Gaussianizing the data after it has been transformed by previous layers. According to our preliminary results, this seems to be the case.
We constructed a two-stage model, trained greedily layer-by-layer, consisting of the transformations CICA--GDN--CICA--GDN. The first CICA instance implements a complete, invertible linear transformation with a set of 256 convolutional filters of support $48 \times 48$, with each filter response subsampled by a factor of 16 (both horizontally and vertically). The output thus consists of 256 reduced-resolution feature maps.  The first GDN operation then acts on the 256-vectors of responses at a given spatial location across all maps. Thus, the responses of the first CICA--GDN stage are Gaussianized across maps, but not across spatial locations. The second-stage CICA instance is applied to vectors of first-stage responses across all maps within a $9 \times 9$ spatial neighborhood -- thus seeking new non-Gaussian directions across spatial locations \emph{and} across maps. Histogram estimates of the marginals of these directions are shown in figure~\ref{fig:multilayer}. The distributions are qualitatively similar to those found for the first stage CICA operating on image pixels, although their heavy-tailedness is less pronounced. The figure also shows histograms of the second-stage GDN marginals, indicating that the new directions have been effectively Gaussianized.

\section{Conclusion}
We have introduced a new probability model for natural images, implicitly defined in terms of an invertible nonlinear transformation that is optimized so as to Gaussianize the data.  This transformation is formed as the composition of a linear operation and a generalized form of divisive normalization, a local gain control operation commonly used to model response properties of sensory neurons.  We developed an efficient algorithm for fitting the parameters of this transformation, minimizing the KL divergence of the distribution of transformed data against a Gaussian target. The resulting density model is not closed-form (because we need to evaluate the determinant of the Jacobian matrix), but it does allow direct computation of probability/likelihood, and is readily used for sampling and inference.

Our parametric transformation includes previous variants of divisive normalization as special cases, and the induced density model generalizes forms of ICA/ISA and elliptically symmetric models.  We show that the additional complexity of our generalized normalization transform allows a significant increase in performance, in terms of Gaussianization, denoising, and sampling.  In addition, we found that the fitted parameters of our model (in particular, the interactions governed by $\bm \gamma$) do not resemble any of these special cases (not shown), and we expect that their detailed structure will be useful in elucidating novel statistical properties of images. It will also be important to compare this induced density model more thoroughly to other model forms that have been proposed in the literature (e.g., finite mixtures of Gaussians or GSMs \citep{GuSiPo08,LySi09,ZoWe12,ThHoBe12}, and sparse factorization \citep{CuSoOl11}).

Our method arises as a natural combination of concepts drawn from two different research endeavors.  The first aims to explain the architecture and functional properties of biological sensory systems as arising from principles of coding efficiency \citep{Ba61,RiBoBi95,BeSe97,ScSi01}.
A common theme in these studies is the idea that the hierarchical organization of the system acts to transform the raw sensory inputs into more compact, and statistically factorized, representations.  Divisive normalization has been proposed as a transformation that contributes to this process. The new form we propose here is highly effective: the transformed data are significantly closer to Gaussian than data transformed by either marginal or radial Gaussianization, and the induced density is thus a more factorized representation of the data.

The second endeavor arises from the statistics literature on projection pursuit, and the use of Gaussianization in problems of density estimation \citep{FrStSc84}. More recent examples include marginal and radial transformations \citep{ChGo00,LySi09a,LaCaMa11}, as well as rectified-linear transformations \citep{DiKrBe14}. Our preliminary experiments indicate that the fusion of a generalized variant of the normalization computation with the iterated Gaussianization architecture is feasible, both in terms of optimization and statistical validity. We believe this architecture offers a promising platform for unsupervised learning of probabilistic structures from data, and are currently investigating techniques to jointly optimize the stages of more deeply stacked models.

\section{Appendix}
\label{sec:appendix}
\subsection{Negentropy}
To see that the negentropy $J$ of the transformed data $\bm y$ can be written as an expectation over the original data, consider a change of variables:
\begin{align*}
J(p_{\bm y}) &= \E_{\bm y} \Bigl( \log p_{\bm y}(\bm y) - \log \mathcal N( \bm y ) \Bigr) \\
&= \int p_{\bm y}(\bm y) \Bigl( \log p_{\bm y}(\bm y) - \log \mathcal N( \bm y ) \Bigr) \D \bm y \\
&= \int p_{\bm x}(\bm x) \Bigl| \frac {\partial \bm y} {\partial \bm x} \Bigr|^{-1} \Biggl( \log \Bigl( p_{\bm x}(\bm x) \Bigl| \frac {\partial \bm y} {\partial \bm x} \Bigr|^{-1} \Bigr) - \log \mathcal N(\bm y) \Biggr) \Bigl| \frac {\partial \bm y} {\partial \bm x} \Bigr| \D \bm x \\
&= \E_{\bm x} \Biggl( \log p_{\bm x}(\bm x) - \log\Bigl| \frac {\partial \bm y} {\partial \bm x} \Bigr| - \log \mathcal N(\bm y) \Biggr)
\end{align*}

\subsection{Invertibility}
Here, we show that a transformation $g: \bm x \mapsto \bm y$ is invertible if it is continuous and its Jacobian $g': \bm x \mapsto \frac {\partial \bm y} {\partial \bm x}$ positive definite everywhere. First note that $g$ is invertible if and only if any two nonidentical inputs $\bm x_a$, $\bm x_b$ are mapped to nonidentical outputs $\bm y_a$, $\bm y_b$, and vice versa:
\begin{equation*}
\forall \bm x_a, \bm x_b: \quad \bm x_a \neq \bm x_b \; \Leftrightarrow \; \bm y_a \neq \bm y_b.
\end{equation*}
Since $g$ is a function, the left-hand inequality follows trivially from the right. To see the converse direction, we can write the inequality of the two right-hand side vectors as
\begin{equation*}
\exists \bm u: \bm u^\T \Delta \bm y \neq 0,
\end{equation*}
where $\Delta \bm y$ is their difference. Second, we can compute $\Delta \bm y$ by integrating the Jacobian along a straight line $L_{ab}$ between $\bm x_a$ and $\bm x_b$:
\begin{equation*}
\Delta \bm y = \int_{L_{ab}} g'(\bm x) \D \bm x.
\end{equation*}
Writing the integral as a Riemann limit, invertibility can be stated as:
\begin{equation*}
\bm x_a \neq \bm x_b \; \Leftrightarrow \; \exists \bm u: \bm u^\T \Delta \bm y = \lim_{T \to \infty} \sum_{t=0}^{T-1} \bm u^\T g'\Bigl(\bm x_a + t\,\frac{\Delta \bm x}{T}\Bigr) \frac {\Delta \bm x} {T} \neq 0.
\end{equation*}
If $\Delta \bm x \neq \bm 0$ (the left-hand inequality is true) and $g'$ is positive definite everywhere, all terms in the sum can be made positive by choosing $\bm u = \frac {\Delta \bm x} {T}$. Hence, for $g'$ positive definite, the right-hand inequality follows from the left.

\subsection{Preprocessing}
We performed two preprocessing steps on the van Hateren dataset before fitting our model: removal of images with saturation artifacts and passing the intensity values through a nonlinearity. To remove heavily saturated images, we computed a histogram of intensity values for each of the images. If more than 0.1\% of the pixel values were contained in the highest-valued histogram bin, we removed the image from the dataset. We passed the remaining 2904 images through a pointwise nonlinearity. In the literature, a logarithm is most commonly used, although there is no particularly convincing reason for using precisely this function. Since the GDN densities are zero-mean by definition, mean removal is necessary to fit the density. Instead of the logarithm, we used the inverse of a generalized logistic function, which is very similar, but can be chosen to marginally Gaussianize the intensity values, which is in line with our objective and also removes the mean.

For the Kodak dataset, we converted the integer RGB values to linear luminance intensities using the transformation specified in the sRGB colorspace definition. Then, a pointwise nonlinearity was fitted to the data to remove the mean and marginally Gaussianize the intensity values, analogous to the nonlinearity used on the van Hateren dataset. To follow established conventions for the denoising experiment, the gamma removal/pointwise nonlinearity was considered part of the model.

\subsubsection*{Acknowledgments}
JB and EPS were supported by the Howard Hughes Medical Institute. VL was supported by the APOSTD/2014/095 Generalitat Valenciana grant (Spain).

\bibliography{main}
\bibliographystyle{iclr2016_conference}

\end{document}